\documentclass[10pt,twocolumn,letterpaper]{article}

\usepackage{cvpr}
\usepackage{times}
\usepackage{epsfig}
\usepackage{subfigure}
\usepackage{booktabs}
\usepackage{multirow}
\usepackage{url}
\usepackage[T1]{fontenc}
\usepackage[utf8]{inputenc}
\usepackage{graphicx}
\usepackage{amsmath,amssymb} 
\usepackage{color}
\usepackage{algorithm}
\usepackage[noend]{algpseudocode}
\usepackage{arydshln}
\usepackage{amsthm}
\usepackage{scrextend}
\usepackage{color,soul}

\def\mb{\mathbf}

\def\eg{\emph{e.g., }}

\def\etc{\emph{etc.}}

\usepackage[pagebackref=true,breaklinks=true,letterpaper=true,colorlinks,bookmarks=false]{hyperref}

\cvprfinalcopy 

\def\cvprPaperID{1221} 

\ifcvprfinal\fi
\begin{document}

\title{Modeling Local Geometric Structure of \\ 3D Point Clouds 
using Geo-CNN}


\author{Shiyi Lan$^{1}$ ~~~Ruichi Yu$^{1}$ 
~~~ Gang Yu$^{2}$
~~~ Larry S. Davis$^1$\\
$^1$University of Maryland, College Park ~~~~~~~~~~ $^2$Megvii Inc (Face++)\\
{\tt\small {sylan@cs.umd.edu, \{yrcbsg,  lsd\}@umiacs.umd.edu},  yugang@megvii.com
}
}

\maketitle

\begin{abstract}
Recent advances in deep convolutional neural networks (CNNs) have motivated researchers to adapt CNNs to directly model points in 3D point clouds. 
Modeling local structure has been proven to be important for the success of convolutional architectures, and researchers exploited the modeling of local point sets in the feature extraction hierarchy. However, limited attention has been paid to explicitly model the geometric structure amongst points in a local region. 
To address this problem, we propose Geo-CNN, which applies a generic convolution-like operation dubbed as GeoConv to each point and its local neighborhood. Local geometric relationships among points are captured when extracting edge features between the center and its neighboring points.
We first decompose the edge feature extraction process onto three orthogonal bases, and then aggregate the extracted features based on the angles between the edge vector and the bases. This encourages the network to preserve the geometric structure in Euclidean space throughout the feature extraction hierarchy. GeoConv is a generic and efficient operation that can be easily integrated into 3D point cloud analysis pipelines for multiple applications. We evaluate Geo-CNN on ModelNet40 and KITTI and achieve state-of-the-art performance.
\end{abstract}

\section{Introduction}
With the development of popular sensors such as RGB-D cameras and LIDAR, 3D point clouds can be easily acquired and directly processed in many computer vision tasks \cite{navigation,shapethetic,fc_pc,pc_registration,segmentation,deformation,3dpose,edge,3D-R2N2,flow}. 
Although hand-crafted features on point clouds have been utilized for many years, recent breakthroughs came with the development of convolutional neural networks (CNNs) inspiring researchers to adapt insights from 2D image analysis with CNNs to point clouds. 

One intuitive idea is to convert irregular point clouds into regular 3D grids by voxelization \cite{volumetric2-voxnet, voxelNet,volumetric1_shapenet,volumetric3}, which enables CNN-like operations to be applied. However, volumetric methods suffer from insufficient resolution, due to sparsely-occupied 3D grids and the exponentially increasing memory cost associated with making the grid finer.
To learn 3D representation at high resolution, kd-tree and octree based methods hierarchically partition space to exploit input sparsity \cite{kd-tree,octree}. But those methods focus more on subdivision of a volume rather than local geometric structure. 
An important architectural model that directly processes point sets is PointNet \cite{pointnet}, which aggregates features from points using a symmetric function. To improve the ability to handle local feature extraction, PointNet++ \cite{pointnet++} aggregates features in local regions hierarchically. However, these methods still ignore the geometric structure amongst points by treating points independently in the global or local point sets. 

\begin{figure}[t]
\centering
  \includegraphics[width=0.9\linewidth]{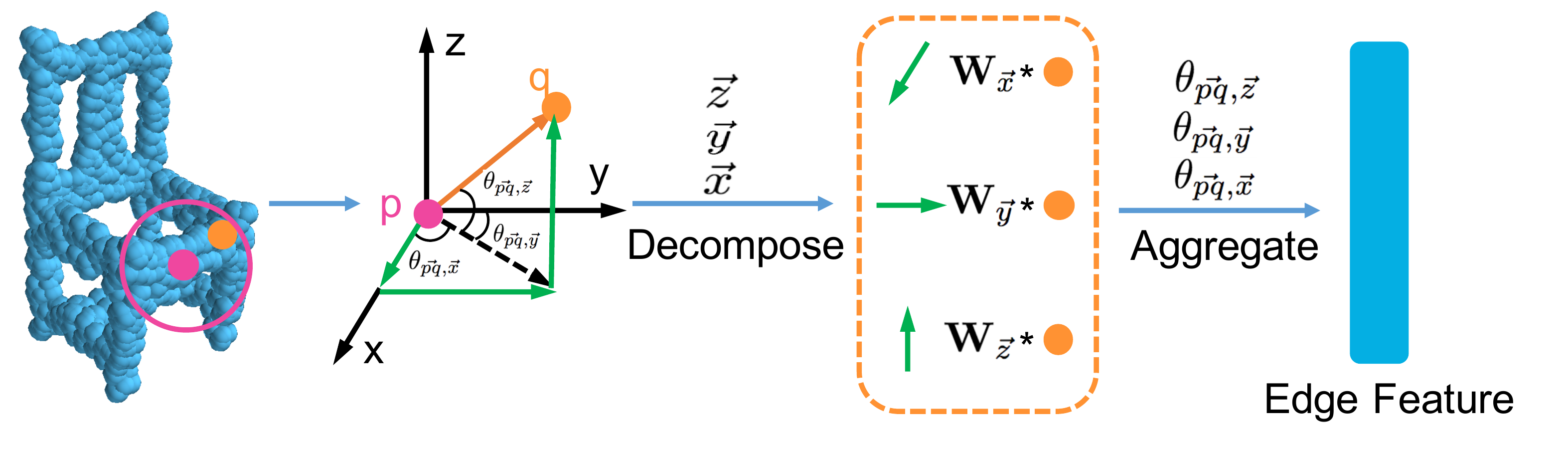}
  \caption{Modeling Geometric Structure between Points via Vector Decomposition. We first decompose the edge features along three orthogonal directions and apply direction-associated weights to extract directional descriptions. Then we aggregate them according to the vector's orientation to construct compact edge features between point $p$ and $q$.
}
\label{fig:edge}
\end{figure}

One recent attempt to model geometric relationships between points is EdgeConv \cite{edgeconv}, which extracts features from each point and its local k-nearest-neighborhood. EdgeConv extracts \textit{edge features} between a center point and its neighboring points. 
The geometric structure between two points $p$ and $q$ is represented by the vector $\vec{pq}$.
However, EdgeConv only models the distance between points (which is the norm of $\vec{pq}$) when constructing the neighborhood, and it ignores the \textit{direction} of the vector, which leads to loss of local geometric information. 
Considering that 3D coordinates are given at the input level of most point cloud analysis pipelines, One might reasonably assume that geometric information might be implicitly learned directly from the coordinates. 
However, current methods may have the following two challenges for geometric modeling: first, the geometric relationship amongst points may be overwhelmed by the large variance of the 3D coordinates, which makes it difficult to be learned from data directly; second, current methods project the 3D coordinates to some high-dimensional space, which may not preserve the geometric structure of the points in the original Euclidean space, especially when the feature extraction hierarchy is deep. 

To address these problems, we propose a novel convolution-like operation \textit{GeoConv} to explicitly model the geometric structure amongst points throughout the hierarchy of feature extraction. GeoConv is applied to each point and its local spherical neighborhood determined by a radius. 
As shown in Fig.\ref{fig:edge}, the vector $\vec{pq}$ which represents the geometric structure between two points, can be decomposed into three orthogonal bases. Based on this vector decomposition, we project the edge features between the two points into three fixed orthogonal bases (the $\vec{x},\vec{y},\vec{z}$ in Fig.\ref{fig:edge}) and apply direction-associated weight matrices ($\mb{W}_{\vec{x}},\mb{W}_{\vec{y}},\mb{W}_{\vec{z}}$ in Fig.\ref{fig:edge}) to extract features along each direction; then we aggregate them proportional to the angle between $\vec{pq}$ and the bases ($\theta_{\vec{pq},\vec{x}},\theta_{\vec{pq},\vec{y}},\theta_{\vec{pq},\vec{z}}$ in Fig.\ref{fig:edge}). 
By decomposing the edge feature extraction process into three orthogonal directions, we reduce the variance of the absolute coordinates of the point cloud, and encourage the network to learn edge features along each basis independently; by aggregating the features according to the geometric relationship between the edge vector and the bases, we explicitly model the geometric structure amongst points. 
Learning in this fashion decomposes the complex geometric structure learning problem into simpler ones while still preserving geometric information. 
Finally, to extract local features of the center point, we weight the edge features from all points in the local neighborhood based on the norm of $\vec{pq}$. 
Another advantage of GeoConv is that it enables feature level multi-view augmentation. Our decomposition-aggregation method enables us to approximate the rotation of point clouds at the feature level via re-weighting the features by manipulating the angles. 


By stacking multiple layers of GeoConv with increasing size of neighborhoods, we construct Geo-CNN, to hierarchically extract features with increasing receptive fields. We aggregate the features from all points by channel-wise max pooling to maintain permutation invariance. Geo-CNN is a generic module that models local geometric structure of points. It can be easily integrated into different pipelines for 3D point cloud analysis, \eg 3D shape classification, segmentation and object detection. We evaluate Geo-CNN on ModelNet40 \cite{volumetric1_shapenet} and KITTI \cite{kitti} and achieve  state-of-the-art performance. 

\section{Related Work}
Motivated by the recent development in 3D sensor technology, increasing attention has been drawn to developing efficient and effective representations on 3D point clouds for shape classification, shape synthesis and modeling, indoor navigation, 3D object detection, \etc \cite{3dgraph,spiderCNN,foldingnet,SPLATNet,rnnslice,upsample,registration,depth,pcvlad,ls,shapecontext,supportspace,ppfnet,psgn}. 
Some earlier works constructed hand-crafted feature descriptors to capture local geometric structure and model local similarity between shapes \cite{hand1,hand2,hand3,hand4,hand5,hand6}. 
More recently, deep neural networks have been used to learn representations directly from data. 
One intuitive way to model the unstructured geometric data is voxelization, which represents a point cloud as a regular 3D grid over which 3D ConvNets can be easily applied \cite{voxelNet,volumn1,volumetric1_shapenet,volumetric2-voxnet,volumetric3,volumn2,volumn3,volumn4}. However, volumetric methods usually produce 3D grids which are sparsely occupied in the space. Their exponentially growing computational cost associated with making the grid finer limits the resolution in each volumetric grid, and leads to quantization artifacts. Due to its regular structures and scalability compared to uniform grids, some indexing techniques such as kd-tree and octree have also been applied to model point clouds \cite{kd-tree,octree}, but those methods still focus more on subdivision of a volume rather than modeling local geometric structure. 

To directly model each 3D point individually, PointNet \cite{pointnet}, PointNet++ \cite{pointnet++} and their variations \cite{frustum,so-net} aggregated point features by a symmetric function to construct a global descriptor. Instead of working on individual points, some recent works exploited local structures by constructing a local neighborhood graph and applying convolution-like operations on the edges connecting neighboring pairs of points \cite{edgeconv}. However, in contrast to our proposed Geo-CNN, all of the above methods do not explicitly model the geometric structure of 3D points, which is represented by the norm and orientation of the vector between two points. Our proposed GeoConv operation models the geometric structure of points by a decomposition and aggregation method based on vector decomposition, and can be easily integrated into different pipelines for 3D object recognition, segmentation and detection tasks \cite{pointnet++,pointnet,frustum,voxelNet}.

Instead of modeling the native 3D format of a point cloud, view-based techniques represent a 3D object as a collection of 2D views, which is compatible with standard CNNs used for image analysis tasks \cite{volumn1,mv1-mvcnn,mv2-rotationNet,mv2_driving,mv-4}. To aggregate information from different orientations of a 3D object, multi-view methods are applied to pool the features extracted from different rendered 2D views, and usually yield better performance than using a single view. Inspired by this, we augment different orientations of the 3D points via approximating rotations of the input point clouds at feature level to further improve the performance of our model.
\section{Our Approach}
We propose a generic operation \textit{GeoConv} to explicitly model the geometric structure in a local region. 
By stacking several layers of GeoConv with increasing receptive field, we construct a Geometric-induced Convolution Neural Network (Geo-CNN) to hierarchically extract features that preserve the geometric relationships among points in Euclidean space. We then aggregate the features from each point by channel-wise max-pooling to extract a global feature descriptor of point clouds.

\subsection{Hierarchical Feature Extraction with Geo-CNN}
\begin{figure*}[t]
\centering
  \includegraphics[width=0.9\linewidth]{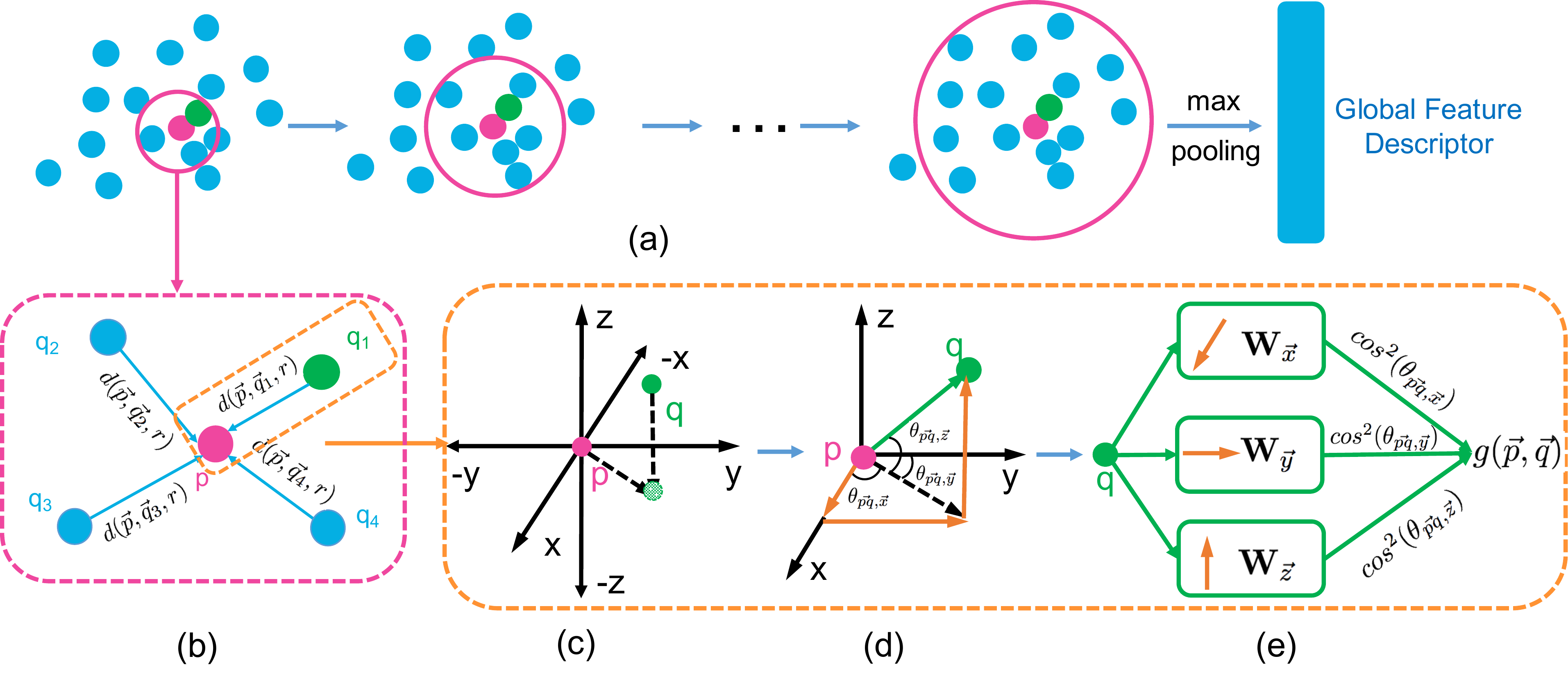}
  \caption{Geo-induced Convolution Neural Network (Geo-CNN). We apply Geo-CNN to hierarchically extract feature representations from a point set. For each point $p$, GeoConv is applied to its local spherical neighborhood defined by a radius $r$. We enlarge the receptive field of GeoConv by increasing $r$ at higher levels of the network (shown as the larger circles in (a)). In the local neighborhood of $p$, we compute the edge features between point $p$ and all neighboring points $q's$, and weight them with a distance measurement function $d(\cdot)$ as shown in (b). To extract the edge features between point $p$ and $q$, we first localize the quadrant that point $q$ belongs to, in a coordinate system with $p$ as its origin, as illustrated in (c). Then,
  we compute the edge features along the three bases of that quadrant by direction-associated weight matrices represented as $\mb{W}_{\vec{x}},\mb{W}_{\vec{y}},\mb{W}_{\vec{z}}$, and aggregate them according to the angles between vector $\vec{pq}$ and the three bases, shown as $\theta_{\vec{pq},\vec{x}},\theta_{\vec{pq},\vec{y}},\theta_{\vec{pq},\vec{z}}$ in (d-e). 
}
\label{fig:GeoConv}
\end{figure*}

With a set of 3D points as input, we exploit local geometric structure by applying a convolutional-like operation (GeoConv) on each point and its local neighborhood. 
We build the Geo-CNN by stacking multiple GeoConv layers with increasing neighborhood size. 
We progressively enlarge the receptive field of the convolution and abstract larger and larger local regions, to hierarchically extract features and preserve the geometric structure of points along the hierarchy (as shown in (a) of Fig.\ref{fig:GeoConv}).

Consider a $C$ dimensional point cloud with $n$ points. We denote the feature of point $p$ at the $l^{th}$ layer of Geo-CNN as $\mb{X}^l_{\vec{p}} \in \mathbb{R}^C$. Usually the 3D points at the input level are represented by their 3D coordinates, but we could also include additional features to represent appearance, surface normal, \etc 
$\,$ For each point $p$, we construct its local neighborhood using a sphere centered at that point with radius $r$. GeoConv is applied on point $p$ and all points $q$ in the neighborhood $N(\vec{p})$, where $N(\vec{p}) = \{ \vec{q} \ | \ \|\vec{p}-\vec{q}\| \leqslant r\}$. The general formula of GeoConv operation applied at the neighborhood of point $p$ at layer $l+1$ is:
\begin{equation}
    \begin{aligned}
        \mb{X}^{l+1}_{\vec{p}} = & s(\vec{p}) + \sum_{\vec{\vec{q}} \in N(\vec{p})}h(\vec{p}, \vec{q}, r)  \\
        = & \mb{W}_c \mb{X}^l_{\vec{p}} + 
        \frac{\sum_{\vec{\vec{q}} \in N(\vec{p})} d(\vec{p},\vec{q},r) g(\vec{p}, \vec{q})  }{\sum_{\vec{q} \in N(\vec{p})} d(\vec{p},\vec{q},r)}
    \end{aligned}
    \label{equa:geo-all}
\end{equation}
where we aggregate features from the center point $\vec{p}$ and the edge features that represent the relationship between the center point and its neighboring points.
$\mb{W}_c$ is the weight matrix used to extract features from the center point.
$g(\vec{p},\vec{q})$ is the function that models edge features, and we weight the features from different neighboring points according to the distance between point $\vec{p}$ and $\vec{q}$ using $d(\vec{p},\vec{q},r)$ as:
\begin{equation}
    \begin{aligned}
        d(\vec{p},\vec{q},r) = (r - \|\vec{p}-\vec{q}\|) ^ 2
    \end{aligned}
    \label{equa:distanceFunc}
\end{equation}
$d(\vec{p},\vec{q},r)$ satisfies two desired properties: (1) monotonically decreasing with $\|\vec{p}-\vec{q}\|$; (2) as $r$ increases, which means as the receptive field of our operation becomes larger, the difference of the weight function $d(\cdot)$ between points that have similar distance to the center point $p$ will decrease. 

After several GeoConv layers, we apply channel-wise max-pooling to aggregate features of each individual point to construct a global feature descriptor of the point cloud. This feature descriptor can be fed into a classifier for 3D shape recognition, segmentation or detection network. GeoConv is a generic operator that can be easily integrated into current 3D point set analysis pipelines to extract local features while preserving geometric structure in Euclidean space.

\subsection{GeoConv: Local Geometric Modeling with Basis-based Decomposition and Aggregation}
The most important part of the GeoConv operation is the way it models edge features.
A straightforward way would be to apply a neural network or multi-layer perceptron (MLP) to compute its activation against each edge. 
However, this method could easily suffer overfitting due to the large variance of edge geometry, which is represented by the vector $\vec{pq}$. 
On the other hand, the above operation may also project the features into some high dimensional space, where the original Euclidean geometric structure among points is not preserved.
In 3D Euclidean space, any vector can be represented by its projection on the three orthogonal bases ($\vec{x}, \vec{y}, \vec{z}$), and the projection norm of the vector on each basis represents the "energy" along that direction.  
So, we decompose the process of edge feature extraction using the three orthogonal bases: we apply direction-associated weight matrices $\mb{W}_{\vec{b}}$ to extract edge features along each direction independently. Then, we aggregate the direction-associated features based on the projection of the vector $\vec{pq}$ on each basis to preserve geometric structure. 
In practice, to differentiate between positive and negative directions of each basis, we consider six bases represented as:
\begin{equation}
    \begin{aligned}
        B = \{& (1, 0, 0), (-1, 0, 0), (0, 1, 0),  \\
                        & (0, -1, 0), (0, 0, 1), (0, 0, -1) \}
    \end{aligned}
    \label{eq:B}
\end{equation}

As shown in (c) of Fig.\ref{fig:GeoConv}, the six bases separate the space into 8 quadrants, and any vector in a specific quadrant can be composed by three bases out of $B$. Given a neighboring point $q$, we first localize the quadrant it lies in (we consider a relative coordinate system by setting $p$ as the origin).
Then we project the vector $\vec{pq}$ onto the three bases of this quadrant, and compute the angle between $\vec{pq}$ and each basis (shown in Fig.\ref{fig:GeoConv} (d)). 
We apply the direction-associated weight matrices represented as $\mb{W}_{\vec{b}}$ to extract the component of edge features along each direction, and aggregate them as shown below:
\begin{equation}
    \begin{aligned}
        g(\vec{p},\vec{q}) = \sum_{\vec{b} \in B_{\vec{q}}} &cos^2(\theta_{\vec{pq},\vec{b}})  \mb{W}_{\vec{b}}  \mb{X}^l_{\vec{q}} 
    \end{aligned}
    \label{equa:geoconv}
\end{equation}
where $\mb{X}^l_{\vec{q}}$ is the feature of point $q$ at the $l^{th}$ layer, and $B_{\vec{q}}$ is a set consisting of three bases selected from $B$ according to the quadrant in which point $\vec{q}$ lies.
The feature along each direction is aggregated with the coefficients $cos^2(\theta_{\vec{pq},\vec{b}})$, which corresponds to the square of the ratio between the norm of each projected component of $\vec{pq}$ and the norm of $\vec{pq}$, and they naturally sum to 1. 

By modeling edge geometry using the basis-based decomposition, our network learns to extract representations for each direction independently. This reduces the complexity of the learning task, when compared with directly learning from the large variance of input 3D coordinates. 
By aggregating the features along each basis, we explicitly model geometric structure of the edge vector between each point and its neighbors.
By learning geometric modeling using GeoConv, we model and preserve the geometric structure of 3D point clouds at every level of our hierarchical feature extraction framework.

\subsection{Approximating 3D Multi-view Augmentation at the Feature Level using Geo-CNN}
Inspired by previous works \cite{mv1-mvcnn,mv2-rotationNet} that aggregate information of a 3D object by utilizing rendered 2D images with different virtual camera views, 
we can also sample from different orientations by rotating 3D points, and then pool the multi-view representations to augment information from different views.
In the 3D space, any rotation of the point clouds can be decomposed into the rotation around the $\vec{z}$ axis and around the plane spanned by $\vec{x}$ and $\vec{y}$. For simplicity, "rotation" in this paper refers to rotation around the $\vec{z}$ axis; our analysis can be easily expanded to other cases.

A naive way to incorporate multiple 3D views at training time is to use the rotated point sets as data augmentation, but this method usually leads to even worse performance in our baseline as well as our implementation of some other works (\eg \cite{pointnet++,pointnet}). A possible reason is that the current methods cannot efficiently learn a compact model from the large variance introduced by multiple 3D views. An alternative is to train a specific model for each 3D view and aggregate the output of multiple networks, which will dramatically increase model complexity.

Instead of input-level multi-view augmentation, we approximate rotation at the feature level in our network using the GeoConv operation. This is done by sharing the computations on edge features along different directions and only changing the aggregation model.
Specifically, we approximate multi-view training and testing by manipulating the aggregation step in GeoConv:
\begin{equation}
    \begin{aligned}
        g_{MV}(\vec{p},\vec{q}) = \sum_{v \in V} w_v  \sum_{\vec{b} \in B_{\vec{q}}} &cos^2(\theta_{\vec{pq}_v,\vec{b}}) \mb{W}_{\vec{b}} \mb{X}_{\vec{q}} 
    \end{aligned}
    \label{equa:geoconv_mv}
\end{equation}
where $w_v$ are learned weights to fuse multi-view features; $\theta_{\vec{pq}_v,\vec{b}}$ are the re-computed angles between the rotated edge vector and the fixed bases.

\section{Implementation Details}
\begin{figure}[t]
\centering
  \includegraphics[width=0.8\linewidth]{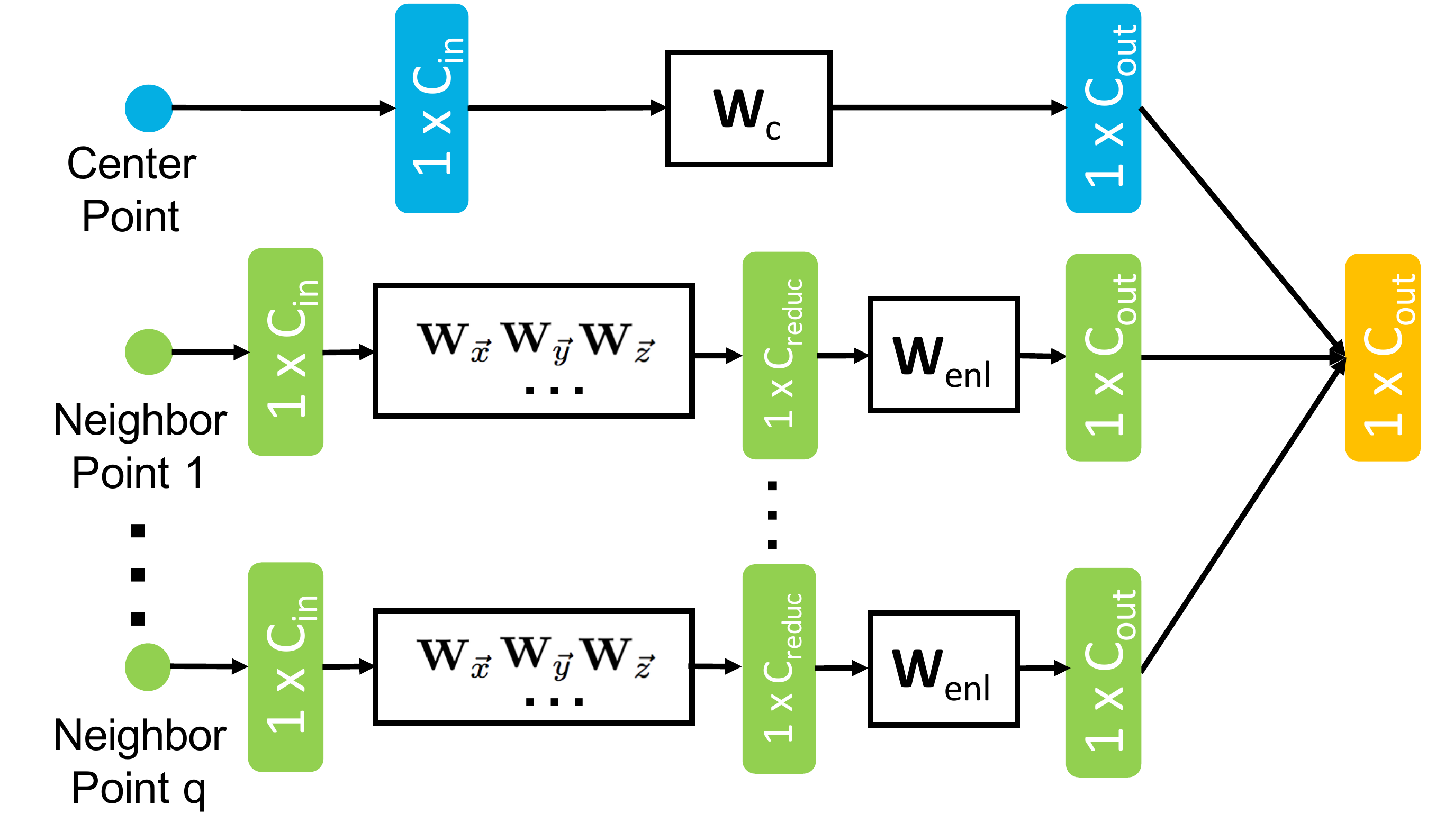}
  \caption{Implementation of GeoConv. The filled boxes show the point features with their dimensionality. The black boxes are the operations. We employ a bottleneck-like structure to first use our decomposition-aggregation method to extract edge features with a lower dimensionality, and then enlarge the dimensionality to match the features extracted from the center point. We aggregate edge features from each point following Eq.\eqref{equa:geo-all}.
}
\label{fig:geoconv_imp}
\end{figure}

\subsection{GeoConv Module}
The input/output of a GeoConv layer is $n \times C_{in}$ and $n \times C_{out}$, where $n$ is the number of points, $C_{in}$ and $C_{out}$ are the input/output dimensionality of each point feature. 
For each point, we construct its local spherical neighborhood defined by a hyper-parameter $r$. 
We apply a weight matrix $\mb{W}_c$ with size $C_{in} \times C_{out}$ to extract features from the center point. For edge feature extraction, we apply a bottleneck module inspired by ResNet\cite{resnet} to first extract features with lower-dimension $C_{reduc}$ (we refer to this layer as "\textit{reduction layer}"), and then enlarge their dimensionality as shown in Fig.\ref{fig:geoconv_imp}. 
The hyper-parameter of GeoConv is the radius $r$. In practice, we split the training data into training and validation set and apply cross-validation to choose the radius for each layer.

\subsection{Geo-CNN for 3D Shape Classification} 
For 3D shape classification on ModelNet40 \cite{volumetric1_shapenet}, we randomly sample 1,000 points from the 3D model of an object. The input features are the 3D coordinates and the surface normal (6 input channels in total). 
Geo-CNN has two branches: (1) similar to PointNet++\cite{pointnet++}, we sample 16 nearest-neighbors from each of the 1,000 points and apply three fully-connected (FC) layers with output dimentionality as 64-128-384 on each group of 16 points. The output size is $1 \times 384$ for each of the 1000 point.
(2) For the second branch, we feed the same input points into an FC layer to project them into a 64-dim feature space. Then we apply the first GeoConv layer with $C_{in}=64$, $C_{reduc}=64$ and $C_{out}=128$. An FC layer with 256 output channels and a GeoConv layer with $C_{in}=256$, $C_{reduc}=64$ and $C_{out}=512$ follow. 
At this point, we channel-wisely concatenate the features extracted from the two branches to obtain an 896-dim feature vector for each point. Next, we apply the third GeoConv with $C_{in}=896$, $C_{reduc}=64$ and $C_{out}=768$ followed by the last FC layer with 2048-dim output. 
Channel-wise max-pooling is then applied to aggregate features from all points. 
We conduct shape classification on the pooled global feature descriptor. The radius for constructing local neighborhoods for the three GeoConv layers are 0.15, 0.3 and 0.6 (the 3D coordinates are normalized in ModelNet40).
Batch-normalization and ReLU are applied after every FC layer and each reduction layer in the GeoConv module.

\subsection{Geo-CNN for 3D Object Detection}
As a generic feature extraction module, Geo-CNN can be easily applied in any pipeline for point-based 3D object detection. 
We follow Frustum PointNet V1 pipeline and replace some layers in the segmentation network with GeoConv layers. 
There are 3 MLP modules in the 3D Instance Segmentation PointNet of Frustum V1 \cite{frustum} with 9 FC layers in total for feature extraction. We directly replace all of the FC layers with the GeoConv layers. For fair comparison, the output dimensionalities of the GeoConv layers are exactly the same as the replaced FC layers. The radii of the 2 layers in the first MLP block are 0.15-0.2; the radii for the 3 layers in the second block are 0.3-0.4-0.4; for the 4 layers in the third block, the radii are 0.3-0.2-0.1-0.1. 
We also explored replacing the FC layers in the box estimation module of Frustum PointNet, but obtained slightly worse results. One possible reason is that when comparing with segmentation, bounding box regression depends more on modeling global information of an object, rather than modeling geometric structures of local point sets. 

For the 3D object detection pipeline, we construct frustums based on 2D box proposals generated by the object detection pipeline (similar to 2D object detection methods \cite{faster,fast,yu1,yu2}) in \cite{frustum}. Then, the Point Segmentation Network with GeoConv is applied to categorize the points on the object in each frustum and eliminate the noise caused by background point clouds. Finally, we use the same Box Estimation Network as \cite{frustum} to obtain the orientation, size and centroid of the 3D bounding box. The implementation of GeoConv is the same as ModelNet. 

\subsection{Baselines} Our baselines have very similar architecture with Geo-CNN, with two difference in the edge feature extraction process: first, baselines fuse the edge features from different points by simply averaging them, unlike GeoConv which weights  the features based on the distance measurement $d(\cdot)$; second, at the reduction layer, GeoConv utilizes three separate weights along each direction, while the baseline applies a single weight matrix to extract edge features.

\section{Experiments}


To demonstrate the effectiveness of Geo-CNN on modeling 3D point clouds, we evaluate it on 3D shape classification using ModelNet40 \cite{volumetric1_shapenet} with CAD-generated points. In addition, a more practical application of 3D point clouds analysis is autonomous driving, which involves 3D point clouds acquired from 3D sensors such as LIDAR. Since GeoConv is a generic operation that can be easily applied in standard pipelines analyzing 3D point clouds, we integrate it into the recent Frustrum PointNet framework to evaluate on 3D object detection using KITTI\cite{kitti}.

\subsection{3D Shape Classification on CAD-Generated 3D Point Clouds}
\subsubsection{Dataset}
We first evaluate our model on the ModelNet40 \cite{volumetric1_shapenet} data with 3D point clouds generated from CAD. There are 12,311 CAD models from 40 object categories, with 9,843 training and 2,468 testing samples.
The CAD models are represented by vertices and faces. For fair comparisons with previous works, we use the prepared ModelNet40 dataset from \cite{pointnet++}, where each model is represented by 10,000 points. One can also sample various sizes of point clouds, e.g., 1000 or 5,000, from the point set.

\subsubsection{Comparison with other Methods}

Table \ref{table:modelnet40} shows the comparison between our Geo-CNN and prior methods.
Geo-CNN achieves state-of-the-art performance on the object classification task with both evaluation metrics of ModelNet40\footnote{Since the two metrics are very similar, "performance" on this dataset refers "Accuracy Overall" metric.}. Our baseline achieves similar performance with the state-of-the-art PointNet++\cite{pointnet++}. By simply changing the operation on modeling the edge features in a local point set from a fully-connected layer to GeoConv, we obtain a gain of $1.6\%$, which demonstrates the effectiveness of our geometrical modeling method.
By further approximating 3D multi-view at the feature level, we obtain a further $0.5\%$ performance gain. We implement the multi-view approximation by virtually rotating the point clouds around the z-axis from 0 to 360 degrees. 
We uniformly approximate 30 views \footnote{The performance gain of multi-view approximation is robust to the number of views from 10 to 40, with less than $\pm$0.1 changes.} following Eq.\eqref{equa:geoconv_mv}. 
Our method, which is applied directly on point clouds, even outperforms single networks with multi-view images, \eg\cite{volumn1} (92\%) and \cite{mv1-mvcnn}(90.1\%), and achieves comparable performance with the method integrated multiple networks in \cite{volumn1} (93.8\%). However, our single model Geo-CNN with approximated multi-view learning at feature level is more scalable and flexible compare to multi-view representations using multiple networks.

\begin{table}[t]
\caption{ModelNet40 Shape Classification Results. We sort the previous methods by time. }
\centering
\small
\begin{tabular}{c|c|c}
\toprule
Method     & \begin{tabular}[c]{@{}c@{}}Accuracy \\Overall\end{tabular} & \begin{tabular}[c]{@{}c@{}}Accuracy \\Class\end{tabular} \\ \hline
PointNet\cite{pointnet}    & 89.2             & 86.2               \\
PointNet++\cite{pointnet++}  & 91.9             & -                  \\
DeepSets\cite{deepsets}    &      90.3        &     -           \\
ECC\cite{ECC}          &    87.4          &       83.2         \\
OctNet\cite{octree}             &   86.5           &     83.8           \\
O-CNN\cite{O-CNN}               &  90.6            &         -       \\
Kd-Net\cite{kd-tree} & 91.8             & 88.5               \\
EdgeConv\cite{edgeconv} & 92.2 & 90.2 \\
SO-Net\cite{so-net}  & 93.4             & 90.8               \\
SpiderCNN\cite{spiderCNN} & 92.4 & -\\
SCN\cite{shapecontext}&90.0 & 87.6 \\
MRTNet\cite{multires_tree}   &     91.7        &       -         \\ 
SpecGCNN\cite{spectureGCNN}  &     92.1        &       -         \\
\hline
Our baseline &  91.8 & 88.2 \\
Geo-CNN    & 93.4             &  91.1         \\ 
Geo-CNN+ MV-Approx. & \textbf{93.9}             & \textbf{91.6}    \\
\bottomrule       
\end{tabular}
\label{table:modelnet40}
\end{table}

\subsection{3D Object Detection on LIDAR Points}

Prior methods on 3D point cloud analysis were mostly evaluated exclusively on artificially generated data. However, with the development of 3D sensing techniques, 3D point clouds can be easily acquired in many real world applications, \eg autonomous driving. The distribution of real point clouds could vary significantly from generated points. For instance, generated data contains dense points from various orientations. However, point clouds obtained by sensors, \eg LIDAR, only contain points from frontal surfaces due to occlusion. Moreover, LIDAR point clouds are noisier and contain a large amount of background, while generated point clouds contain pure on-object points. Evaluation on real data such as point clouds collected by LIDAR is very important to show the robustness and practicality of a 3D point cloud analysis method. 
To illustrate the effectiveness of Geo-CNN on real-world 3D points, we evaluate on 3D object detection using the KITTI dataset \cite{kitti}.

\begin{table*}[h]
\centering
\begin{tabular}{c|ccc|ccc|ccc}
\toprule
\multirow{2}{*}{Method} & \multicolumn{3}{c|}{Cars} & \multicolumn{3}{c|}{Pedestrians} & \multicolumn{3}{c}{Cyclists} \\ \cline{2-10} 
                        & Easy   & Moderate & Hard   & Easy     & Moderate    & Hard     & Easy    & Moderate   & Hard    \\ \hline
VoxelNet\cite{voxelNet}                & 81.97  & 65.46    & 62.85  & 57.86    & 53.42       & 48.87    & 67.17   & 47.65      & 45.11   \\
Frustum PointNet v1\cite{frustum}     &   83.33     &   69.00       &    61.97    &   71.65       &    53.43         &   49.20      &   67.29      &    56.74        &     49.84            \\
Frustum PointNet v2\cite{frustum}     & 83.42  & 70.40    & 63.37  & 70.51    & 55.31       & 52.11   & 64.30   & 57.33      & 50.43   \\ \hline
Baseline  &    84.56    &   69.16       &  62.50      &    72.32      &  51.36           &   47.70      &  64.68      &   55.59         &    48.45    \\
Frustum Geo-CNN                  & \textbf{85.09}  & \textbf{71.02}    & \textbf{63.38} & \textbf{75.64}    & \textbf{56.25}       & \textbf{52.54}   & \textbf{69.64}   & \textbf{60.50}      & \textbf{52.88}  \\
\bottomrule
\end{tabular}
\label{table:kitti}
\caption{Performance Comparison in 3D Object Detection: average precision (in \%) on KITTI validation set. Geo-CNN achieves significantly better performance when compared with the baseline, which demonstrates the effectiveness of our decomposition-aggregation method on modeling local geometry. Our Frustum Geo-CNN is implemented based on Frustum PointNet v1, and it outperforms both Frustum PointNet v1 and v2.}
\end{table*}


\subsubsection{Dataset}
The KITTI 3D object detection benchmark contains 7,481 training and 7,518 testing images/point clouds with three object categories: \textit{Car}, \textit{Pedestrian}, \textit{Cyclist}. For each class, detection outcomes are evaluated based on three difficulty levels: easy, moderate, and hard. The level of difficulty is based on object size, occlusion state, and truncation level. 
For fair comparison with the state-of-the-art detection methods, we directly replace the PointNet feature extraction module in the Frustum PointNet v1 \cite{frustum} detection pipeline with Geo-CNN, and use the 2D bounding box proposals released by \cite{frustum} in our experiments. Since only train/val proposals of frustum pointnet are published, we conduct evaluation using the protocol described in \cite{frustum,voxelNet} and use their training/testing split. 

\subsubsection{Comparison with other Methods}
\begin{figure*}[t]
\centering
  \includegraphics[width=0.8\linewidth]{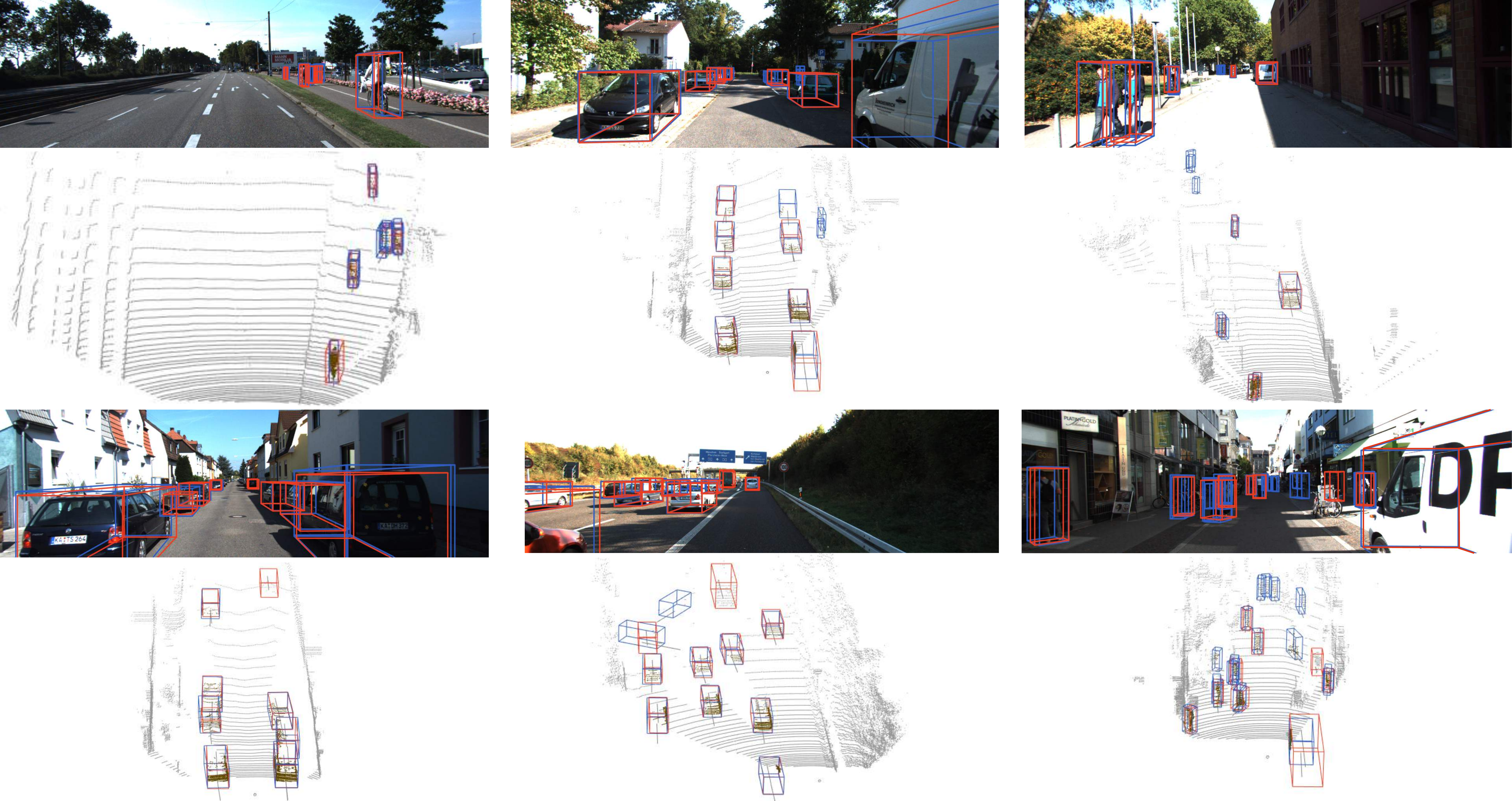}
  \caption{We visualize the detection results on KITTI with 2D and 3D images. The red boxes are the groundtruth boxes and the blue boxes are the prediction results. Some of the false positive detection results are because of missing annotation.
}
\label{fig:v}
\end{figure*}

Table \ref{table:kitti} shows the evaluation results on KITTI 3D object detection. 
Our implementation of the detection pipeline is based on Frustum PointNet v1, which involves object proposals in 2D object detection \cite{faster,fast,yu3,yu4,yu5}. The performance of v1 was surpassed by Frustum PointNet v2, which has a more complex architecture. However, by replacing the PointNet feature extraction module in the segmentation network of v1 with GeoConv, Frustum with Geo-CNN outperforms both Frustum PointNet v1 and v2. The performance of Frustum v1 and v2 on the validation set is evaluated based on the released code of \cite{frustum}, and it is very similar with the performance reported in \cite{frustum}.
We visualize the detection results of Frustum with Geo-CNN on 2D and 3D images in Fig.\ref{fig:v}.

\subsection{Ablation Study}
\noindent \textbf{Can we model local geometry by directly learning from 3D coordinates?}
We study different ways to model local geometry between points in the feature extraction hierarchy. 
Since the geometric structure is encoded in the 3D coordinates of points, one straightforward way to learn geometric structure is to apply an FC layer to directly learn from the coordinates. However, previous hierarchical feature extraction methods project the 3D coordinates input to some high-dimensional feature space at the first layer of their networks, which may lead to the loss of the Euclidean geometry amongst points. 
Our baseline method takes the 3D coordinates at the input level to directly learn the geometric structure implicitly.
To preserve the Euclidean geometry throughout the feature extraction hierarchy, we apply an FC layer to learn the geometric structure between points directly from the 3D coordinates of point $p$ and $q$ at \textit{every layer} of our baseline model, and concatenate the extracted features with the original ones channelwisely. We refer to this method as "Baseline + 3D Coords". 
We also investigated alternative approaches to model the angle of the vector $\vec{pq}$ in GeoConv. Instead of using $g(\cdot)$ function as proposed, we directly learn these aggregation coefficients using an FC layer with the 3D coordinates of point $p$ and $q$ as input. We refer to this method as "GeoConv - Learned-Agg". 
As shown in Table \ref{table:ablation_xyz}, directly learning geometric structure between points or the coefficients to aggregate the decomposed features from the 3D coordinates does not help. This reveals that modeling the local geometry is non-trivial, and GeoConv effectively captures geometric structures amongst points to improve the feature extraction framework.


\begin{table}[]
\centering
\begin{tabular}{@{}c|c@{}}
                      & Accuracy Overall \\ \midrule
Baseline              &   91.8          \\ 
Baseline + 3D Coords  & 91.7             \\
GeoConv - Learned-Agg & 91.5            \\
GeoConv               & \textbf{93.4}             \\ 
\end{tabular}
\caption{\textbf{Ablation Study: Different Geometric Modeling Methods}. We study different ways to model local geometry amongst points using ModelNet40 dataset. "Baseline + 3D Coords" directly learns the geometric structure with 3D coordinates of the two points at \textit{every layer} of the network; "GeoConv - Learned-Agg" aggregates the direction-associated features by learned weights.  }
\label{table:ablation_xyz}
\end{table}




\bigskip
\noindent \textbf{Does the performance gain of GeoConv come from increasing model complexity?}
By decomposing the edge feature extraction process into three directions using separate neural networks, GeoConv increases the number of parameters in the reduction layer of edge feature extraction.
The number of parameters for the edge feature extraction operation is $C_{in} * n_{bases} * C_{reduc} + C_{reduc} * C_{out}$, where $C_{in}$ and $C_{out}$ are the input/output channel numbers. $c_{reduc}$ is the output channel number of the channel reduction step, and the difference between GeoConv and the baseline is $n_{bases}$ ($n_{bases}=6$ for GeoConv and $n_{bases}=1$ for the baseline).
We increase $c_{reduc}$ from 64-64-64 to 192-192-256 for the three reduction layers in the baseline model to roughly match the number of parameters for edge feature extraction of GeoConv operation. 
The enlarged baseline is referred to as "Baseline-Large" shown in Table \ref{table:abl_param}, which is evaluated on the ModelNet40 classification task. 
It is worth noting that the number of parameters in the reduction layers accounts for a very small portion of the total number of parameters, and the experiment setting of the other components of the networks except for the reduction layers, are exactly the same for both baseline and Geo-CNN.
It is clear that simply enlarging the number of channels does not improve performance, and the gain of GeoConv is not due to the increasing number of parameters.

\begin{table}[]
\centering
\footnotesize
\begin{tabular}{c|ccc}
Method   & Baseline & Baseline-Large & GeoConv \\ \hline
Accuracy Overall &  91.8    & 91.7           & \textbf{93.4}    \\\hline
\begin{tabular}[c]{@{}c@{}}\#. of parameters for\\edge feature extraction\end{tabular} & 167.9K & 610.8K & 557.1K
\end{tabular}
\\
\label{table:abl_param}
\caption{\textbf{Ablation Study: Model Complexity}. We add channels to the weight matrix of the reduction layer of the baseline method (Baseline-Large) to match the number of parameters of GeoConv. We show the sum of number of parameters of the 3 reduction layers in each model. 
The results on ModelNet40 shows that simply increasing model complexity does not help.}
\end{table}

\begin{table}[]
\centering
\begin{tabular}{c|c}
Method                    & Accuracy Overall \\ \hline
Baseline                  & 91.8            \\
Baseline + Data Aug.       & 91.6             \\
Geo-CNN                   & 93.4             \\
Geo-CNN + Data Aug.       & 92.6             \\ \hline
Geo-CNN + MV-Approx. & \textbf{93.9}             \\
\end{tabular}
\label{table:abl_aug}
\caption{Overall Accuracy on ModelNet40 with Different Multi-view Augmentations. "Data Aug." and "MV-Approx." refer to input-level augmentation and our feature-level multi-view approximation.}
\end{table}


\bigskip
\noindent \textbf{3D Multi-view Augmentation.}
We evaluate the effect of our feature-level multi-view augmentation. As a straightforward way to incorporate multi-view information to the network learning process, one can simply rotate the input point clouds randomly as data augmentation at training time. On the other hand, our proposed decomposition-aggregation method in GeoConv enables us to approximate 3D multi-view augmentation at the feature level. 
Table \ref{table:abl_aug} shows the performance of input-level multi-view augmentation and feature-level approximation on ModelNet40 dataset. We observe that input-level multi-view data augmentation leads to performance degradation of both the baseline method and Geo-CNN. One possible reason is that the input-level data augmentation brings in large variance between different views, which cannot be properly learned with a single compact model. Another possible solution is to learn separate models with different views and then aggregate them. However, the models with multiple networks are less flexible and scalable due to their high complexity.

\section{Conclusion}
We address the problem of modeling local geometric structure amongst points with GeoConv operation and a hierarchical feature extraction framework dubbed Geo-CNN. Inspired by the success of exploiting local structure using CNNs on 2D image analysis task, we propose to extract features from each point and its local neighborhood with a convolutional-like operation. GeoConv explicitly models the geometric structure between two points by decomposing the feature extraction process onto three orthogonal directions, and aggregating the features based on the angles between the edge vector and the bases. The Geo-CNN with GeoConv operation achieves state-of-the-art performance on the challenging ModelNet40 and KITTI datasets.

\section*{Acknowledgement}
The research was partially supported by the Office of Naval Research under Grant N000141612713: Visual Common Sense Reasoning for Multi-agent Activity Prediction and Recognition.


{\small
\bibliographystyle{ieee}
\bibliography{egbib}

\begin{thebibliography}{10}\itemsep=-1pt

\bibitem{pcvlad}
M.~Angelina~Uy and G.~Hee~Lee.
\newblock Pointnetvlad: Deep point cloud based retrieval for large-scale place
  recognition.
\newblock In {\em The IEEE Conference on Computer Vision and Pattern
  Recognition (CVPR)}, June 2018.

\bibitem{hand1}
M.~Aubry, U.~Schlickewei, and D.~Cremers.
\newblock The wave kernel signature: {A} quantum mechanical approach to shape
  analysis.
\newblock In {\em {ICCV} Workshops}, pages 1626--1633. {IEEE}, 2011.

\bibitem{hand3}
R.~Beserra~Gomes, B.~M. Ferreira~da Silva, L.~K. d.~M. Rocha, R.~V. Aroca,
  L.~C. P.~R. Velho, and L.~M.~G. Gon\c{c}alves.
\newblock Efficient 3d object recognition using foveated point clouds.
\newblock {\em Comput. Graph.}, 37(5):496--508, Aug. 2013.

\bibitem{volumn2}
A.~Brock, T.~Lim, J.~Ritchie, and N.~Weston.
\newblock Generative and discriminative voxel modeling with convolutional
  neural networks.
\newblock pages 1--9, 12 2016.
\newblock Workshop contribution; Neural Inofrmation Processing Conference : 3D
  Deep Learning, NIPS ; Conference date: 05-12-2016 Through 10-12-2016.

\bibitem{volumetric3}
A.~Brock, T.~Lim, J.~M. Ritchie, and N.~Weston.
\newblock Generative and discriminative voxel modeling with convolutional
  neural networks, 2016.
\newblock cite arxiv:1608.04236Comment: 9 pages, 5 figures, 2 tables.

\bibitem{hand2}
M.~M. Bronstein and I.~Kokkinos.
\newblock Scale-invariant heat kernel signatures for non-rigid shape
  recognition.
\newblock In {\em CVPR}, pages 1704--1711. IEEE Computer Society, 2010.

\bibitem{mv2_driving}
X.~Chen, H.~Ma, J.~Wan, B.~Li, and T.~Xia.
\newblock Multi-view 3d object detection network for autonomous driving.
\newblock In {\em IEEE CVPR}, 2017.

\bibitem{3D-R2N2}
C.~B. Choy, D.~Xu, J.~Gwak, K.~Chen, and S.~Savarese.
\newblock 3d-r2n2: A unified approach for single and multi-view 3d object
  reconstruction.
\newblock In {\em Proceedings of the European Conference on Computer Vision
  ({ECCV})}, 2016.

\bibitem{ppfnet}
H.~Deng, T.~Birdal, and S.~Ilic.
\newblock Ppfnet: Global context aware local features for robust 3d point
  matching.
\newblock In {\em The IEEE Conference on Computer Vision and Pattern
  Recognition (CVPR)}, June 2018.

\bibitem{volumn3}
M.~Engelcke, D.~Rao, D.~Z. Wang, C.~H. Tong, and I.~Posner.
\newblock Vote3deep: Fast object detection in 3d point clouds using efficient
  convolutional neural networks.
\newblock In {\em {ICRA}}, pages 1355--1361. {IEEE}, 2017.

\bibitem{psgn}
H.~Fan, H.~Su, and L.~J. Guibas.
\newblock A point set generation network for 3d object reconstruction from a
  single image.
\newblock In {\em 2017 {IEEE} Conference on Computer Vision and Pattern
  Recognition, {CVPR} 2017, Honolulu, HI, USA, July 21-26, 2017}, pages
  2463--2471, 2017.

\bibitem{multires_tree}
M.~Gadelha, R.~Wang, and S.~Maji.
\newblock Multiresolution tree networks for 3d point cloud processing.
\newblock In {\em The European Conference on Computer Vision (ECCV)}, September
  2018.

\bibitem{yu2}
M.~Gao, A.~Li, R.~Yu, V.~I. Morariu, and L.~S. Davis.
\newblock C-wsl: Count-guided weakly supervised localization.
\newblock {\em arXiv preprint arXiv:1711.05282}, 2017.

\bibitem{yu1}
M.~Gao, R.~Yu, A.~Li, V.~I. Morariu, and L.~S. Davis.
\newblock Dynamic zoom-in network for fast object detection in large images.
\newblock {\em IEEE Conference on Computer Vision and Pattern Recognition
  (CVPR)}, 2018.

\bibitem{3dpose}
L.~Ge, Z.~Ren, and J.~Yuan.
\newblock Point-to-point regression pointnet for 3d hand pose estimation.
\newblock In {\em The European Conference on Computer Vision (ECCV)}, September
  2018.

\bibitem{kitti}
A.~Geiger.
\newblock Are we ready for autonomous driving? the kitti vision benchmark
  suite.
\newblock In {\em Proceedings of the 2012 IEEE Conference on Computer Vision
  and Pattern Recognition (CVPR)}, CVPR '12, pages 3354--3361, Washington, DC,
  USA, 2012. IEEE Computer Society.

\bibitem{fast}
R.~Girshick.
\newblock Fast {R-CNN}.
\newblock In {\em Proceedings of the International Conference on Computer
  Vision ({ICCV})}, 2015.

\bibitem{shapethetic}
A.~Golovinskiy, V.~G. Kim, and T.~Funkhouser.
\newblock Shape-based recognition of {3D} point clouds in urban environments.
\newblock {\em International Conference on Computer Vision (ICCV)}, Sept. 2009.

\bibitem{hand4}
Y.~Guo, M.~Bennamoun, F.~A. Sohel, M.~Lu, and J.~Wan.
\newblock 3d object recognition in cluttered scenes with local surface
  features: {A} survey.
\newblock {\em {IEEE} Trans. Pattern Anal. Mach. Intell.}, 36(11):2270--2287,
  2014.

\bibitem{resnet}
K.~He, X.~Zhang, S.~Ren, and J.~Sun.
\newblock Deep residual learning for image recognition.
\newblock In {\em {CVPR}}, pages 770--778. {IEEE} Computer Society, 2016.

\bibitem{rnnslice}
Q.~Huang, W.~Wang, and U.~Neumann.
\newblock Recurrent slice networks for 3d segmentation of point clouds.
\newblock In {\em The IEEE Conference on Computer Vision and Pattern
  Recognition (CVPR)}, June 2018.

\bibitem{mv2-rotationNet}
A.~Kanezaki, Y.~Matsushita, and Y.~Nishida.
\newblock Multi-view convolutional neural networks for 3d shape recognition.
\newblock In {\em Proc. CVPR}, 2018.

\bibitem{kd-tree}
R.~Klokov and V.~S. Lempitsky.
\newblock Escape from cells: Deep kd-networks for the recognition of 3d point
  cloud models.
\newblock In {\em ICCV}, 2017.

\bibitem{ls}
L.~Landrieu and M.~Simonovsky.
\newblock Large-scale point cloud semantic segmentation with superpoint graphs.
\newblock In {\em The IEEE Conference on Computer Vision and Pattern
  Recognition (CVPR)}, June 2018.

\bibitem{yu4}
A.~Li, J.~Sun, J.~Y.-H. Ng, R.~Yu, V.~I. Morariu, and L.~S. Davis.
\newblock Generating holistic 3d scene abstractions for text-based image
  retrieval.
\newblock {\em IEEE Conference on Computer Vision and Pattern Recognition
  (CVPR)}, 2017.

\bibitem{so-net}
J.~Li, B.~M. Chen, and G.~Hee~Lee.
\newblock So-net: Self-organizing network for point cloud analysis.
\newblock In {\em The IEEE Conference on Computer Vision and Pattern
  Recognition (CVPR)}, June 2018.

\bibitem{deformation}
K.~Li, T.~Pham, H.~Zhan, and I.~Reid.
\newblock Efficient dense point cloud object reconstruction using deformation
  vector fields.
\newblock In {\em The European Conference on Computer Vision (ECCV)}, September
  2018.

\bibitem{volumn4}
Y.~Li, S.~Pirk, H.~Su, C.~R. Qi, and L.~J. Guibas.
\newblock Fpnn: Field probing neural networks for 3d data.
\newblock {\em CoRR}, abs/1605.06240, 2016.

\bibitem{pc_registration}
Y.~Liu, C.~Wang, Z.~Song, and M.~Wang.
\newblock Efficient global point cloud registration by matching rotation
  invariant features through translation search.
\newblock In {\em The European Conference on Computer Vision (ECCV)}, September
  2018.

\bibitem{volumetric2-voxnet}
D.~Maturana and S.~Scherer.
\newblock Voxnet: A 3d convolutional neural network for real-time object
  recognition.
\newblock In {\em IEEE/RSJ International Conference on Intelligent Robots and
  Systems}, page 922 – 928, September 2015.

\bibitem{frustum}
C.~R. Qi, W.~Liu, C.~Wu, H.~Su, and L.~J. Guibas.
\newblock Frustum pointnets for 3d object detection from rgb-d data.
\newblock In {\em The IEEE Conference on Computer Vision and Pattern
  Recognition (CVPR)}, June 2018.

\bibitem{pointnet}
C.~R. Qi, H.~Su, K.~Mo, and L.~J. Guibas.
\newblock Pointnet: Deep learning on point sets for 3d classification and
  segmentation.
\newblock In {\em {CVPR}}, pages 77--85. {IEEE} Computer Society, 2017.

\bibitem{volumn1}
C.~R. Qi, H.~Su, M.~Nie{\ss}ner, A.~Dai, M.~Yan, and L.~Guibas.
\newblock Volumetric and multi-view cnns for object classification on 3d data.
\newblock In {\em Proc. Computer Vision and Pattern Recognition (CVPR), IEEE},
  2016.

\bibitem{pointnet++}
C.~R. Qi, L.~Yi, H.~Su, and L.~J. Guibas.
\newblock Pointnet++: Deep hierarchical feature learning on point sets in a
  metric space.
\newblock In I.~Guyon, U.~V. Luxburg, S.~Bengio, H.~Wallach, R.~Fergus,
  S.~Vishwanathan, and R.~Garnett, editors, {\em Advances in Neural Information
  Processing Systems 30}, pages 5099--5108. Curran Associates, Inc., 2017.

\bibitem{faster}
S.~Ren, K.~He, R.~Girshick, and J.~Sun.
\newblock Faster r-cnn: Towards real-time object detection with region proposal
  networks.
\newblock In C.~Cortes, N.~D. Lawrence, D.~D. Lee, M.~Sugiyama, and R.~Garnett,
  editors, {\em Advances in Neural Information Processing Systems 28}, pages
  91--99. Curran Associates, Inc., 2015.

\bibitem{supportspace}
Z.~Ren and E.~B. Sudderth.
\newblock 3d object detection with latent support surfaces.
\newblock In {\em The IEEE Conference on Computer Vision and Pattern
  Recognition (CVPR)}, June 2018.

\bibitem{fc_pc}
D.~Rethage, J.~Wald, J.~Sturm, N.~Navab, and F.~Tombari.
\newblock Fully-convolutional point networks for large-scale point clouds.
\newblock In {\em The European Conference on Computer Vision (ECCV)}, September
  2018.

\bibitem{octree}
G.~Riegler, A.~O. Ulusoy, and A.~Geiger.
\newblock Octnet: Learning deep 3d representations at high resolutions.
\newblock In {\em CVPR}, 2017.

\bibitem{depth}
R.~Roveri, L.~Rahmann, C.~Oztireli, and M.~Gross.
\newblock A network architecture for point cloud classification via automatic
  depth images generation.
\newblock In {\em The IEEE Conference on Computer Vision and Pattern
  Recognition (CVPR)}, June 2018.

\bibitem{hand6}
R.~M. Rustamov.
\newblock Laplace-beltrami eigenfunctions for deformation invariant shape
  representation.
\newblock In {\em Proceedings of the Fifth Eurographics Symposium on Geometry
  Processing}, SGP '07, pages 225--233, Aire-la-Ville, Switzerland,
  Switzerland, 2007. Eurographics Association.

\bibitem{hand5}
R.~B. Rusu, N.~Blodow, and M.~Beetz.
\newblock Fast point feature histograms (fpfh) for 3d registration.
\newblock In {\em Proceedings of the 2009 IEEE International Conference on
  Robotics and Automation}, ICRA'09, pages 1848--1853, Piscataway, NJ, USA,
  2009. IEEE Press.

\bibitem{ECC}
M.~Simonovsky and N.~Komodakis.
\newblock Dynamic edge-conditioned filters in convolutional neural networks on
  graphs.
\newblock {\em CoRR}, abs/1704.02901, 2017.

\bibitem{SPLATNet}
H.~Su, V.~Jampani, D.~Sun, S.~Maji, E.~Kalogerakis, M.-H. Yang, and J.~Kautz.
\newblock Splatnet: Sparse lattice networks for point cloud processing.
\newblock In {\em The IEEE Conference on Computer Vision and Pattern
  Recognition (CVPR)}, June 2018.

\bibitem{mv1-mvcnn}
H.~Su, S.~Maji, E.~Kalogerakis, and E.~G. Learned{-}Miller.
\newblock Multi-view convolutional neural networks for 3d shape recognition.
\newblock In {\em Proc. ICCV}, 2015.

\bibitem{registration}
J.~Vongkulbhisal, B.~Irastorza~Ugalde, F.~De~la Torre, and J.~P. Costeira.
\newblock Inverse composition discriminative optimization for point cloud
  registration.
\newblock In {\em The IEEE Conference on Computer Vision and Pattern
  Recognition (CVPR)}, June 2018.

\bibitem{3dgraph}
C.~Wang, B.~Samari, and K.~Siddiqi.
\newblock Local spectral graph convolution for point set feature learning.
\newblock In {\em The European Conference on Computer Vision (ECCV)}, September
  2018.

\bibitem{spectureGCNN}
C.~Wang, B.~Samari, and K.~Siddiqi.
\newblock Local spectral graph convolution for point set feature learning.
\newblock In {\em The European Conference on Computer Vision (ECCV)}, September
  2018.

\bibitem{O-CNN}
P.-S. Wang, Y.~Liu, Y.-X. Guo, C.-Y. Sun, and X.~Tong.
\newblock O-cnn: Octree-based convolutional neural networks for 3d shape
  analysis.
\newblock {\em ACM Trans. Graph.}, 36(4):72:1--72:11, July 2017.

\bibitem{edgeconv}
Y.~Wang, Y.~Sun, Z.~Liu, S.~E. Sarma, M.~M. Bronstein, and J.~M. Solomon.
\newblock Dynamic graph cnn for learning on point clouds.
\newblock {\em arXiv preprint arXiv:1801.07829}, 2018.

\bibitem{volumetric1_shapenet}
Z.~Wu, S.~Song, A.~Khosla, F.~Yu, L.~Zhang, X.~Tang, and J.~Xiao.
\newblock 3d shapenets: A deep representation for volumetric shapes.
\newblock In {\em CVPR}, pages 1912--1920. IEEE Computer Society, 2015.

\bibitem{shapecontext}
S.~Xie, S.~Liu, Z.~Chen, and Z.~Tu.
\newblock Attentional shapecontextnet for point cloud recognition.
\newblock In {\em The IEEE Conference on Computer Vision and Pattern
  Recognition (CVPR)}, June 2018.

\bibitem{spiderCNN}
Y.~Xu, T.~Fan, M.~Xu, L.~Zeng, and Y.~Qiao.
\newblock Spidercnn: Deep learning on point sets with parameterized
  convolutional filters.
\newblock In {\em The European Conference on Computer Vision (ECCV)}, September
  2018.

\bibitem{foldingnet}
Y.~Yang, C.~Feng, Y.~Shen, and D.~Tian.
\newblock Foldingnet: Point cloud auto-encoder via deep grid deformation.
\newblock In {\em The IEEE Conference on Computer Vision and Pattern
  Recognition (CVPR)}, June 2018.

\bibitem{segmentation}
X.~Ye, J.~Li, H.~Huang, L.~Du, and X.~Zhang.
\newblock 3d recurrent neural networks with context fusion for point cloud
  semantic segmentation.
\newblock In {\em The European Conference on Computer Vision (ECCV)}, September
  2018.

\bibitem{edge}
L.~Yu, X.~Li, C.-W. Fu, D.~Cohen-Or, and P.-A. Heng.
\newblock Ec-net: an edge-aware point set consolidation network.
\newblock In {\em The European Conference on Computer Vision (ECCV)}, September
  2018.

\bibitem{upsample}
L.~Yu, X.~Li, C.-W. Fu, D.~Cohen-Or, and P.-A. Heng.
\newblock Pu-net: Point cloud upsampling network.
\newblock In {\em The IEEE Conference on Computer Vision and Pattern
  Recognition (CVPR)}, June 2018.

\bibitem{yu5}
R.~Yu, X.~Chen, V.~I. Morariu, and L.~S. Davis.
\newblock The role of context selection in object detection.
\newblock In {\em British Machine Vision Conference (BMVC)}, 2016.

\bibitem{yu3}
R.~Yu, A.~Li, V.~I. Morariu, and L.~S. Davis.
\newblock Visual relationship detection with internal and external linguistic
  knowledge distillation.
\newblock {\em IEEE International Conference on Computer Vision (ICCV)}, 2017.

\bibitem{mv-4}
T.~Yu, J.~Meng, and J.~Yuan.
\newblock Multi-view harmonized bilinear network for 3d object recognition.
\newblock In {\em The IEEE Conference on Computer Vision and Pattern
  Recognition (CVPR)}, June 2018.

\bibitem{flow}
M.~E. Yumer and N.~J. Mitra.
\newblock Learning semantic deformation flows with 3d convolutional networks.
\newblock In {\em European Conference on Computer Vision (ECCV 2016)},
  pages~--. Springer, 2016.

\bibitem{deepsets}
M.~Zaheer, S.~Kottur, S.~Ravanbakhsh, B.~Poczos, R.~R. Salakhutdinov, and A.~J.
  Smola.
\newblock Deep sets.
\newblock In I.~Guyon, U.~V. Luxburg, S.~Bengio, H.~Wallach, R.~Fergus,
  S.~Vishwanathan, and R.~Garnett, editors, {\em Advances in Neural Information
  Processing Systems 30}, pages 3391--3401. Curran Associates, Inc., 2017.

\bibitem{voxelNet}
Y.~Zhou and O.~Tuzel.
\newblock Voxelnet: End-to-end learning for point cloud based 3d object
  detection.
\newblock In {\em The IEEE Conference on Computer Vision and Pattern
  Recognition (CVPR)}, June 2018.

\bibitem{navigation}
Y.~Zhu, R.~Mottaghi, E.~Kolve, J.~J. Lim, A.~Gupta, L.~Fei-Fei, and A.~Farhadi.
\newblock {Target-driven Visual Navigation in Indoor Scenes using Deep
  Reinforcement Learning}.
\newblock In {\em {IEEE International Conference on Robotics and Automation}},
  2017.

\end{thebibliography}
}

\end{document}